\documentclass[a4paper]{article}
\usepackage{INTERSPEECH2021}
\usepackage{subcaption}
\usepackage{subfiles}

\title{Accented Speech Recognition: A Survey}

\name{
Arthur Hinsvark$^1$,
Natalie Delworth$^1$,
Miguel Del Rio$^1$, 
Quinten McNamara$^1$,
Joshua Dong$^1$,
Ryan Westerman$^1$,
Michelle Huang$^1$,
Joseph Palakapilly$^1$,
Jennifer Drexler$^1$,
Ilya Pirkin$^1$,
Nishchal Bhandari$^1$,
Miguel Jett\'e$^1$
}
\address{
  $^1$Rev.com}
\email{\{
arthur,
natalie.delworth,
miguel.delrio,
quinn,
joshua.dong,
ryan.westerman,
michellehuang,
joseph.palakapilly,
jennifer.drexler,
ilya.pirkin,
nishchal,
miguel \}
@rev.com
}

\begin{document}

\maketitle

\begin{abstract}
Automatic Speech Recognition (ASR) systems generalize poorly on accented speech. The phonetic and linguistic variability of accents present hard challenges for ASR systems today in both data collection and modeling strategies. The resulting bias in ASR performance across accents comes at a cost to both users and providers of ASR.

We present a survey of current promising approaches to accented speech recognition and highlight the key challenges in the space. Approaches mostly focus on single model generalization and accent feature engineering. Among the challenges, lack of a standard benchmark makes research and comparison especially difficult.

  

\end{abstract}

\section{Introduction}
Speech recognition for in-domain audio has reached high levels of performance\cite{xiong2016achieving} and widespread use.
However, ASR performance degrades significantly on accented speech\cite{koenecke2020racial}.
Thus, accent-robustness is needed for speech recognition to be solved in the wild. Generalizing speech recognition across dialects is a hard problem for real-world speech systems.

Dialects are variations within a language that differ in geographical regions and social groups, which can be distinguished by traits of grammar, vocabulary\cite{holmes2013introduction} and, of particular interest to ASR, phonology\cite{yang2018joint}. The terms ``accent'' and ``dialect'' are used interchangeably in the context of speech recognition as both represent the primary speaker-specific phonetic variation in a given language\cite{huang2001analysis}.

Accurate speech transcript data is hard to source, with accurately labeled and transcribed accented speech even more so. This lack of data means standard approaches to speech recognition perform poorly on accented speech. A lack of common training and test suites in the research community makes it difficult to compare similar approaches.

This paper reviews the current state of the art of accent-robust ASR.  We first discuss accent-specific models and how their difficulty in production caused their decline in recent research.
The primary approaches we observed attempt to improve the data-efficiency and generalization capability of modeling from little available training data using techniques such as multi-task training or semi-supervised learning. Feature engineering approaches also showed success, such as direct lexicon modification, accent embeddings, and adversarial training for feature extraction. Finally, there is some research into application of existing ASR systems to minimize error on accents through techniques such as accent-error correction systems.

Sections 2 through 5 showcase the techniques applied in current literature, while Section 6 and 7 discuss the open problems and suggest key areas for future research.






\section{Accent-Specific Modeling}
A straightforward approach to accent-robust ASR design is to use an accent identification system to manage multiple accent-specific ASR systems\cite{najafian2014unsupervised}. However, accent-specific performance depends heavily on the availability of data for the specific accent,
which can be alleviated through data augmentation such as speed modification\cite{fukuda2018data}.
The naive approach is to create a single model, pooling all the data, but this is shown to under-perform compared to accent-specific models\cite{vergyri2010automatic}.
However, training multiple accent-specific models is cumbersome in production and increases maintenance costs. These costs motivate techniques for adapting a unified model that performs well on all accents. 

\section{Model Generalization}
\subsection{Multi-task Learning}
Multi-task learning incorporates additional information into neural network training.
When models trained to perform two (or more) separate tasks share some of their parameters, those shared layers learn to represent the information necessary for all tasks. For ASR, multi-task learning can be used in the context of neural network acoustic models or end-to-end ASR models.
The first use of multi-task learning for accented ASR was when neural network acoustic models first became popular\cite{huang2014multi}; multi-task learning was a natural way to switch from having separate acoustic models for different accents to having accent-specific top-layers but sharing all other acoustic model parameters across accents\cite{huang2014multi,chen2015improving}. While this is both more efficient and better-performing than separate models, it is still a somewhat indirect way of incorporating accent information. 

Prasad and Jyothi\cite{prasad2020accents} presented a variety of in-depth approaches to analyzing DeepSpeech2 neural layers for accented speech.
The authors show that the strongest gradients of the final neural ASR layer for a given word align with the timing interval for that word. The effect is especially strong in US and Canadian accented speech, and weaker in Indian accented speech.
Prasad and Jyothi also show that the first RNN layer in DeepSpeech2 contains the most accent information as demonstrated by the mutual information between the ASR layers and the accent classification. They posit that the preliminary layers of the network are most useful for accent 
classification tasks.

To make full use of accent labels, accent identification is the natural choice for an auxiliary task within a multi-task learning framework. In Yang et al.\cite{yang2018joint}, the authors train an acoustic model with separate output layers for American and British English, as well as an accent classifier that branches off from the first layer of the acoustic model. The authors introduce a tuning parameter, $\alpha$, that scales the accent ID objective relative to the acoustic model objective and is tuned on the development data.

Having separate output layers for different accents requires making a choice of which output layer to use at inference time, based on either knowledge of the input accent or on the output of an accent classifier. Jain et al.\cite{jain2018improved} use a similar approach to Yang et al.\cite{yang2018joint}, but instead use a single output layer for all accents. Additionally, Jain et al.\cite{jain2018improved} explores the use of a stand-alone accent classifier to produce features that are then used as additional input to the ASR model. Combining accent embedding input and multi-task learning performs best.

Li et al.\cite{li2018multi} apply multi-task learning with accent ID to end-to-end ASR. They insert special accent ID tokens to the text labels sequences during training. For example, the output targets for a British accented speech utterance of “hello” would be “\textless sos\textgreater\ \textless en-gb\textgreater\ h e l l o \textless eos\textgreater” instead of “\textless sos\textgreater\ h e l l o \textless eos\textgreater”. This technique makes accent ID part of the ASR objective function and thus does not require any tuning of the trade-off between the ASR objective and the accent ID objective. The authors experiment with different positions for these special tokens and find that they work best at the end. 

Multi-task learning approaches are also useful without accent labels in training data. Rao et al.\cite{rao2017multi} train a single DNN acoustic model with both phonemic and graphemic output targets, and then use the graphemic output targets for inference. The idea behind this technique is that the use of graphemic targets may allow the model to compensate for places where the lexicon does not accurately reflect the accents in the training or test data. At the same time, the authors recognize that graphemic output targets alone perform worse than phonemes, and thus want their model to be able to take advantage of the information contained in a pronunciation dictionary. Multi-task learning allows them to combine both sources of information in one model. 

Ghorbani et al.\cite{ghorbani2018leveraging} explore a special case of multi-task learning to improve recognition of non-native speech. The authors train a multi-lingual end-to-end model with separate output layers for English, Spanish, and Hindi. They then test the English output layer with English speech from native Spanish and Hindi speakers. They show that this model better represents the speech of these non-native English speakers than a model trained on English alone. 

\subsection{Domain Expansion}
Domain expansion is a method for generalizing a model to other domains.  In accented ASR, this method takes a baseline model that performs well on an initial set of accents and improves it's performance on a new accent while maintaining performance on the original set.  Contrast this to fine tuning, where we train on a new accent to improve performance on that accent, but run into the problem of catastrophic forgetting where performance on the original accents decreases dramatically.  Domain expansion is focused on techniques such as regularization and novel architectures to minimize this loss.

Regularization in this context adds a term to the loss that penalizes diverging from the baseline model.  This can be done through KL-divergence\cite{huang2014multi} between the adapted model and the baseline model, elastic weight consolidation (EWC), and weight constraint adaptation (WCA). Ghorbani et al.\cite{ghorbani2019domain} investigates several of these methods for one transfer step finding a hybrid of soft-KL-divergence and EWC to be the best, while Houston\cite{houston2020continual} looks at the performance over multiple transfer steps finding that EWC covers 65\% of the gap between a fine-tuned and a pooled (best-case) model across all accents.  An advantage of these regularization techniques is that the original data need not be present in order to adapt and improve the model.  Architectures techniques such as Progressive Neural Networks\footnote{arXiv:1606.04671} and Progress and Compress\cite{schwarz2018progress} have yet to be applied to this space.

Recently Winata et al.\cite{winata2019learning} applied model agnostic meta learning (MAML) to accented ASR, showing that a model could generalize to unseen accents.  Meta learning, or learning to learn, has also been applied to train models to quickly adapt to new accents in zero or few shots, possibly even at inference\footnote{arXiv:2004.05439}.  However other methods, such as optimizers and different representations, have yet to be applied.

\subsection{Pronunciation Modification}
Many traditional ASR techniques require a pronunciation lexicon, but pronunciation dictionaries are constructed for particular canonical dialects (e.g. American English) which can lead to issues with recognizing accented speech\cite{humphries1998use}. Many pronunciation modification techniques involve learning or discovering a mapping between accented and canonical phones. These mappings may be purely handcrafted and based on domain knowledge\cite{lehr2014discriminative}, learned through training on accented and non-accented speech\cite{goronzy2004generating, loots2011automatic}, or learned given some handcrafted constraints\cite{humphries1996using, arsikere2019multi}. The phone mappings may be used to either augment the pronunciation lexicon\cite{lehr2014discriminative, loots2011automatic}, applied during inference so that a canonical pronunciation lexicon may still be used\cite{goronzy2004generating, arsikere2019multi}, or used as part of a modified, dialect-specific grapheme-to-phoneme model\cite{loots2011automatic}. 

Pronunciation modification techniques are particularly useful for low-resource or no-resource data. \cite{goronzy2004generating} achieves lower WER for a majority of accented speakers using a phone mapping technique and no accented training data. African American Vernacular English (AAVE) may not necessarily be a low resource dialect, but with an off-the-shelf ASR model trained on broadcast news data, Lehr et al.\cite{lehr2014discriminative} uses an augmented lexicon along with LM parameter estimation for a 2.1\% absolute WER gain on AAVE. Humphries et al.\cite{humphries1998use} achieves 9.7\% absolute WER gain on American English using pronunciation modification techniques with an acoustic model trained on British English. When more data is available, however, these techniques may be less useful compared to an accent-specific model - an acoustic model trained on American English achieves 19.3\% absolute WER gain compared to the original British English acoustic model in \cite{humphries1998use}. 

\section{Accent Features}
\subsection{Adaptation}
\subsubsection{Auxiliary Features}
Inspired from successes in speaker adaptation research, a common approach to adapting ASR to accented speech is incorporating accent information into a single generic model. In this section we focus on approaches that augment the input acoustic features with accent information.

After testing out several complex interpolation-of-bases adaptation techniques, Grace et al.\cite{grace2018occam} found that simply adapting the bias of the first layer (mathematically equivalent to concatenating a one-hot encoding of the dialect to the feature vector) performed best and even outperformed dialect-specific models. The conclusion contradicts prior work but the authors claim that the architecture progression (from DNN to LSTM) combined with the large dataset of training audio nullify the need for complex adaptation techniques. Li et al.\cite{li2018multi} and Zhu et al.\cite{zhu2020multi} reached similar conclusions, finding improvement when feeding in a one-hot dialect vector into each component of an LAS architecture or a single hidden layer of a hybrid ASR acoustic model using a gate mechanism, respectively. \cite{li2018multi} was also able to surpass dialect-specific model accuracy, while \cite{zhu2020multi} did not, and saw equally-promising results using a multi-task network with a gate mechanism (a more practical approach without need for accent information a priori).

Yoo et al.\cite{yoo2019highly} build upon this one-hot approach by further integrating dialect information using Feature-wise Linear Modulation (FiLM) to apply feature-wise affine transformations between intermediate LSTM layers conditioned on dialect. The authors find the best results when conditioning on the dialect and an utterance-level feature extractor built on previous layer outputs. \cite{yoo2019highly} sets itself apart through measuring modeling unseen dialects by training an "unknown" accent label from random samples of other dialects. \cite{jain2019multi} echoes a similar philosophy to FiLM, using the Mixture of Experts (MoE) approach to modulate acoustic model input with affine transformations of "experts", covering phonetic variability and accent variability in subsequent expert groups.

\subsubsection{Accent Embeddings}
In recent years, it has become common (inspired by x-vectors in speaker identification literature) to achieve adaptation by augmenting input features with accent embeddings, or vectors extracted from a late hidden layer of an accent identification neural network. Providing embeddings seems to improve both hybrid architecture acoustic models\cite{jain2018improved, turan2020achieving} as well as end-to-end models\cite{viglino2019end, rao2020improved}. Furthermore, the accent embeddings can also be combined with semi-supervised lattice-free maximum mutual information (LF-MMI) fine-tuning\cite{turan2020achieving} and multi-task training with accent identification\cite{jain2018improved, rao2020improved} to improve upon baseline modest gains from using the embeddings alone. Using accent embeddings combined with any of the other approaches cited here seems to be one of the most promising areas for accent adaptation research.

\subsection{Adversarial Training}
Adversarial networks are a popular approach to generative models in computer vision, but they are also useful in speech recognition to improve model robustness to noise\cite{Shinohara2016} and other perturbations. Two papers have recently used adversarial training techniques to improve recognition of accented speech. Sun et al.\cite{Sun2018DomainAT} take inspiration from Domain Adversarial Training (DAT)\cite{Ganin2015UnsupervisedDA} to learn domain-invariant features and minimize the difference between standard and accented speech. Chen et al.\cite{Chen2020AipnetGA} devise AIPNet, a GAN architecture to separate accent-invariant information from audio before passing it along to a typical ASR network.

A DAT\cite{Sun2018DomainAT} architecture consists of three components: the feature generation network; the domain classification network, which discriminates between source and target domains during training; and the senone classification network, which predicts senones during training and inference. In order to learn domain-invariant features, the feature generation network should produce an output feature that makes the domain classification network fail to determine the correct domain, while also remaining  discriminative enough for the senone classifier to identify the correct senone. The loss function for the DAT network contains a hyperparameter $\lambda$, which can be used to switch the operation of the network between DAT and multi-task learning. The authors report that multi-task learning shows inconsistent improvement compared to DAT, and suggest that this is due to the former incorporating domain-discriminative features. DAT has a relative Character Error Rate improvement of 3.8\% on average compared to the baseline trained only on unaccented Mandarin.

AIPNet\cite{Chen2020AipnetGA} consists of a pre-training stage and a fine-tuning stage. During pre-training, the authors use two generators: one accent-specific and one accent-invariant, along with their respective accent discriminators, plus a decoder that is trained to reconstruct the input audio. The authors utilize three loss functions for the pre-training stage: an adversarial loss to train the discriminators, a reconstruction loss to ensure acoustic information is encoded in both generators, and a consistency loss to encourage linguistic information to appear in the accent-invariant generator. Once pre-training is complete, the accent-invariant generator can be connected as an input to any downstream speech task. One major advantage of this technique is that pre-training only needs to be performed once, and can then be used as the base for any number of traditional ASR architectures.

In a supervised setting where transcriptions are available for all accents, AIPNet gives a relative WER improvement of $3.33\%$ on accented speech compared to the baseline. When transcripts are available only for US accented speech, the relative improvement is $6.29\%$. Qualitative analysis of AIPNet shows that the accent-independent generator succeeds in disentangling accent content from linguistic content.

\section{System Combination}

Khandelwal et al.\cite{khandelwal2020black} propose a novel coupling of an open-source accent-tuned local model with a commercial ASR model trained on non-accented speech. The authors introduce the ``FineMerge'' algorithm to enable a commercial ASR system to guide inference of an accent-tuned ASR system at a character level. ``FineMerge'' aligns characters of the commercial transcript with the original audio frames using per-frame character confidence distributions calculated by the open-source decoder. The algorithm then adjusts the confidence of mismatching characters. The authors report that the combined system reduces the commercial system's WER by 17-28\% relative.

\section{Open Problems and Challenges}

\begin{table}

    \textbf{Datasets in Papers Surveyed}
    \centering

    \begin{tabular}{|c | c c c |} 
    \hline
 Paper & Accents & Total Hours & Source \\
     \hline
     \cite{li2018multi,grace2018occam} & 6 & 40000 & Google \\
     \hline
     \cite{arsikere2019multi} & 4 & 40000 & Amazon \\
     \hline
     \cite{rao2017multi} & 4 & 13600 & Google\\
     \hline
     \cite{zhu2020multi} & 4 & 7000 & IoA CAS\\ 
     \hline
     \cite{Chen2020AipnetGA} & 9 & 3800 & Facebook\\
     \hline
     \cite{huang2014multi} & 3 & 2000 & Microsoft \\ 
     \hline
     \cite{houston2020continual} & 4 & 1900 & Amazon \\
     \hline
     \cite{Sun2018DomainAT} & 6 & 960 & NPU, china \\
     \hline
     \cite{jain2019multi} & 8 & 736 & Speech Ocean* \\
     \hline
     \cite{yoo2019highly} & 8 & 500 & Speech Ocean* \\
     \hline
      \cite{vergyri2010automatic} & 6 & 450 & News\\
     \hline
     \cite{chen2015improving} & 6 & 400 & Intel\\
     \hline
     \cite{ghorbani2018leveraging,ghorbani2019domain} & 4 & 200 & UT-CRSS-4\\
     \hline
     \cite{khandelwal2020black} & 3 & 152 & Common Voice*\\
     \hline
     \cite{turan2020achieving} & 5 & 125 & Verbmobil/Voxforge*\\
     \hline
     \cite{najafian2014unsupervised} & 14 & 70 & ABI*\\
     \hline
     \cite{fukuda2018data} & 2 & 43 & Google\\
     \hline
     \cite{shor2019personalizing} & 2 & 37 & ALS\\
     \hline
     \cite{jain2018improved,viglino2019end} & 7 & 34 & Common Voice*\\
     \hline
     \cite{huang2001analysis,huang2004accent} & 4 & 16  & Microsoft\\ 
     \hline
    \end{tabular}
    \caption{Overview of datasets used in accented speech benchmarking. Public datasets denoted by * include Phondat Verbmobil, Voxforge, Mozilla Common voice v4 (MCV-v4)\cite{commonvoice:2020}, and Accents of the British Isles (ABI)}
    \label{tab:survey_data}
\end{table}


\textbf{Standard benchmark}
ASR research in accented speech lacks a standard benchmark, such as ImageNet for image classification. The lack of standardization makes it hard to track progress and compare findings across papers, even when similar techniques are used.

The Artie Bias dataset\cite{meyer2020artie}, based off of the Common Voice Corpus, is a good candidate for such a standard. Artie Bias is a manually verified subset of Common Voice with 17 English accents. However, the sparsity of Chinese, Middle Eastern, European, South and Central American, African, and Australian accents makes it difficult to use as a global benchmark. The dataset is dominated by American, Indian, and British English. Additionally, the test set only uses country labels and lacks finer regional dialect classifications. 



\textbf{Language modeling} A majority of the works surveyed focus on pronunciation differences between dialects, when they can vary in grammar, vocabulary, and spelling as well \cite{holmes2013introduction}.  End-to-end models may capture this implicitly, but systems that use language models for decoding could benefit from research into adapting them to accents.

\textbf{Zero-shot accent robustness} 
More research needs to be done on how ASR systems behave and can adapt to unseen accents. The difficulty of collecting accented English, sparsity of accented speech data, and dynamic nature of language is a major obstacle for general ASR systems in the wild. 
Effectively handling unseen accents could entail moving away from discrete accent categorization and towards more continuous representations of accent attributes. Accent-invariant representations presented in Section 4.2
are promising, though none of the papers in that section discussed unseen accents. Meta-learning may also present a solution for quickly adapting to unseen accents or learning accent-invariant adaptations.

\textbf{Large numbers of accents}
The papers surveyed focus on relatively few accents compared to the entire range seen in the wild, as demonstrated in Table  \ref{tab:survey_data}.
Notably, the ABI dataset shows how a single national accent includes many regional dialects. It is unclear if techniques presented in current literature are able to generalize to the hundreds of dialects and sub-classifications of accents worldwide.

\textbf{Accents in other Languages} English is the most studied language in ASR literature, with much sparser literature for other languages. It is unclear if accent-tuning approaches appropriate for English are applicable to other languages, especially in languages with greater dialect variation such as Hindi, Chinese, or Arabic.

\section{Conclusions}
Accent-robust ASR is increasingly important in practical applications. Significant challenges such as data sparsity and a lack of a standard benchmark impede research progress, and many open questions exist. The research here also shows great progress in the field and a general trend towards successful single-model generalization. We hope this survey serves as a primer for accent-robust ASR research and initiates future research.

\section{Acknowledgements}
We'd like to thank Petr Schwarz of the Brno University of Technology for his help reviewing this survey, and the rest of the team at Rev.com for their feedback on this paper.

\bibliographystyle{IEEEtran}

\bibliography{bib}

\end{document}